\documentclass{article}
\usepackage{spconf,amsmath,graphicx}
\usepackage{xcolor}
\usepackage{hyperref}
\hypersetup{
    colorlinks=true,
    linkcolor=blue,
    filecolor=magenta,      
    urlcolor=blue,
}

\title{BanglaCoNER: Towards Robust Bangla Complex Named Entity Recognition}
\name{HAZ Sameen Shahgir$^{\S ,1}$
    Ramisa Alam$^{\S,1}$
    Md. Zarif Ul Alam$^{\S,1}$
    \thanks{$^{\S}$ Equal Contribution. 
    The soruce code is available at \href{https://github.com/ramisa2108/Bangla-Complex-Named-Entity-Recognition-Challenge}{GitHub}} 
    }
\address{\textsuperscript{\rm 1} Department of CSE, BUET \quad}
\begin{document}
%
\maketitle
\begin{abstract}
Named Entity Recognition (NER) is a fundamental task in natural language processing that involves identifying and classifying named entities in text. But much work hasn’t been done for complex named entity recognition in Bangla, despite being the seventh most spoken language globally. CNER is a more challenging task than traditional NER as it involves identifying and classifying complex and compound entities, which are not common in Bangla language. In this paper, we present the winning solution of Bangla Complex Named Entity Recognition Challenge - addressing the CNER task on BanglaCoNER dataset using two different approaches, namely Conditional Random Fields (CRF) and finetuning transformer based Deep Learning models such as BanglaBERT.

The dataset consisted of 15300 sentences for training and 800 sentences for validation, in the .conll format. Exploratory Data Analysis (EDA) on the dataset revealed that the dataset had 7 different NER tags, with notable presence of English words, suggesting that the dataset is synthetic and likely a product of translation.

We experimented with a variety of feature combinations including Part of Speech (POS) tags, word suffixes, Gazetteers, and cluster information from embeddings, while also finetuning the BanglaBERT (large) model for NER. We found that not all linguistic patterns are immediately apparent or even intuitive to humans, which is why Deep Learning based models has proved to be the more effective model in NLP, including CNER task. Our fine tuned BanglaBERT (large) model achieves an F1 Score of 0.79 on the validation set. Overall, our study highlights the importance of Bangla Complex Named Entity Recognition, particularly in the context of synthetic datasets. Our findings also demonstrate the efficacy of Deep Learning models such as BanglaBERT for NER in Bangla language.
\end{abstract}
\section{Introduction}
\label{sec:intro}

Named Entity Recognition (NER) is a fundamental task in natural language processing that involves identifying and classifying named entities in text. Named entities are words or phrases that refer to specific entities such as persons, locations, organizations, dates, etc. NER is useful for many downstream applications such as information extraction, question answering, summarization, and sentiment analysis.

However, NER is not a trivial task as it requires dealing with various challenges such as ambiguity, variation, and complexity of named entities. Ambiguity refers to the problem of resolving different meanings of the same word or phrase in different contexts. Variation refers to the problem of handling different forms or spellings of the same entity. Complexity refers to the problem of identifying and classifying complex and compound entities that consist of more than one word or phrase.

Complex Named Entity Recognition (CNER) is a more challenging task than traditional NER as it involves identifying and classifying complex and compound entities. Complex entities are those that have internal structure or hierarchy such as nested or overlapping entities. Compound entities are those that are composed of two or more simple entities such as person names with titles or affiliations.

Bangla is the seventh most spoken language in the world with about 250 million speakers. However, much work hasn’t been done for complex named entity recognition in Bangla, despite its linguistic richness and diversity. Bangla has a complex morphology and syntax that pose difficulties for NER systems. Moreover, Bangla has a large vocabulary and a high degree of variation and ambiguity in named entity expressions.

In this paper, we address the CNER task on BanglaCoNER dataset using two different approaches: Conditional Random Fields (CRF) and finetuning transformer based Deep Learning models such as BanglaBERT. CRF is a widely used probabilistic graphical model for sequence labeling tasks such as NER. We experimented with various feature combinations including Part of Speech (POS) tags, word suffixes, Gazetteers, and cluster information from embeddings to improve the performance of CRF. BanglaBERT is a pretrained language model based on BERT architecture that is specifically designed for Bangla language understanding tasks. We fine-tuned BanglaBERT on the CNER dataset to leverage its contextualized representations for NER.

\section{Data Analysis}
\label{sec:data_analysis}

The provided dataset was a labeled Bangla dataset in the \.conll format where each word had a corresponding NER tag and sentences were separated with empty lines.

The train set consists of 15300 sentences and the validation set has 800 sentences. The length of the sentences in both sets varies from 2 words to 35 words with the average length being 12 words.

There are 7 different NER tags in the given dataset with LOC (3804), GRP (6653), PROD (5152), CW (5001), CORP (5299), PER(6738), and O (170K)

There were some English words in the English alphabet and many more English words  transliterated into Bangla, suggestion that the dataset is synthetic and likely a product of translation.

Details EDA can be found in the EDA notebook of the GitHub repository.

\section{Feature Based Learning}
\label{sec:feature_based_learning}

Before the advent of deep learning, feature engineering was a crucial step in building machine learning models, particularly for natural language processing tasks such as named entity recognition. Traditional machine learning algorithms rely on a fixed set of input features to make predictions. These algorithms cannot automatically learn useful representations from raw data, as deep learning models can.

\subsection{Conditional Random Fields}

Conditional Random Fields (CRF) is a type of machine learning algorithm that is commonly used for sequential labeling tasks, such as named entity recognition, where the goal is to assign a label to each word in a sentence. \cite{ekbal2008named} CRF works by modeling the probability of a label sequence given the input feature sequence.

In CRF, feature engineering is important because it directly affects the quality of the input features, which in turn affects the performance of the model. Features that are relevant to the task and provide useful information about the input data will lead to better performance of the CRF model.

Let $x = {x_1, x_2, ..., x_n}$ be the input feature sequence, where $x_i$ is the feature vector of the i-th word.
Let $y = {y_1, y_2, ..., y_n}$ be the corresponding label sequence, where $y_i$ is the label of the i-th word.
Let $\Phi(x, y)$ be a feature function that maps the input feature sequence and the label sequence to a real-valued feature vector.
Let $w$ be the weight vector that parameterizes the feature function.

The goal of CRF is to find the most likely label sequence given the input feature sequence, which can be formulated as:

$$
\arg\max_{y} P(y|x) = \arg\max_{y}\left(\frac{exp(w^T\Phi(x,y))}{\sum_{y'} exp(w^T\Phi(x,y'))}\right)
$$

\subsection{Feature Engineering}

We developed our system by experimenting with a variety of feature combinations. The goal of selecting these features was to improve the ability of our classifier to generalize to new data and handle large amounts of data.

It is important to note that choosing features for Bangla Language turned out to be somewhat challenging compared to English. This is because some of the features that work well for English can't be used for Bangla. For example, it makes sense to check if a word is a title because it could be Person (PER), Location (LOC) etc. But, there is no such concept in Bangla. So, we had to choose the features carefully. 

The details of the collection of features used for the NER task are provided below.

\begin{itemize}
\item \textbf{Part of Speech (POS)}: POS tags can provide useful information about the grammatical structure of the sentence and the role of a word in it. POS tags can be used as a feature in CRF to help disambiguate named entities from other types of words. Here we have used the Bengali POS Tag model, which is a CRF based POS tagger. The POS tagger was trained with nltr dataset with 80\% accuracy. For each word, we have used the POS tag information for that word and $k$ words before and after it. We have experimented with several values of $k$, and we concluded that $k>3$ seems to overfit. So, we used $k = 2$ in the final model for generating features.

\item \textbf{Word suffix}: Word Suffix is an important feature in case of Bangla NER. For example, locations (LOC) can end with suffixes "aloy", "pur", "bari", "liya", "para". Similarly, most female names in Bangla end with "a". We leverage the suffix information for this kind of case. And our experiment shows that the macro F1 score increases by a significant margin in our validation set.

\item \textbf{Word prefix}: Word Prefix is similarly important information for NER. We used a fixed-length prefix of the current and/or the surrounding words.

\item \textbf{k neighbour words}: Neighbouring words can carry useful information about NER of a word. 
We added the previous and next $k$ words of the current word as a feature for different values of $k$. Based on the validation results, we used $k = 2$ in the final model for generating features.

\item \textbf{Cluster Information from Embeddings}: Instead of using Word Embeddings directly, we took a slightly different approach. As the dataset is diverse and there are domain shifts, as well as the out-of-vocabulary words, we didn't think it would be a good approach to use them directly. And if we were to do that, the model won't be robust to unseen data. Our intuition for this feature was that when we meet unknown data, it will be closer to cluster where another NER of its kind belongs.

So, we went for the Word Clustering \cite{mishra2016semi} approach. Our algorithm is as follows - 

\begin{enumerate}
    \item Collect and preprocess a large dataset of Bangla text for training the NER model.
    \item Train a word embedding model on the dataset, such as word2vec, GloVe, or fastText. The embeddings will capture the semantic and syntactic information of the words in the text.
    \item Use the trained embedding model to generate embeddings for each word in the dataset.
    \item Apply clustering techniques, such as k-means or hierarchical clustering, on the word embeddings to group similar words together.
    \item Train the CRF NER model on the dataset, using the word embeddings and cluster information as features.
\end{enumerate}

Here, we have used word2vec for extracting the word vectors and used k-means for clustering. We have also experimented with soft clustering using Gaussian Mixture Models (GMM), but because of the runtime overhead, we discarded that.

\item \textbf{Gazetteer Lists using Bangla t5 NMT}: We used the idea of Gazetteer Lists to further improve our model. The problem was that there were no Gazetteer lists for Bangla to get started. We could have created that with a semi-supervised approach, but considering the domain and diversity of our dataset we took a different approach. 
We used the Bangla t5 NMT model \cite{hasan-etal-2020-low} developed by CSE BUET to create a new set of Bangla Gazetteers. The reasoning was that our dataset had lots of words like "New York", "Czecho-Slovakia", "Arizona". If we were to curate Bangla Gazetteers only using Bangla data, we would not be able to get this kind of data. The CSE BUET Bangla t5 model which was fine-tuned for Neural Machine Translation (NMT) task excels at this kind of task. When given names of places and creative words it is able to generate proper transliteration instead of just translating it to a Bangla word that doesn't make sense. We had score improvements in the creative word category using this method.

\item \textbf{Digit} We also tried using digit information as a feature. But from our experiment, we saw that the impact was not significant.

\end{itemize}

\section{Deep Learning Model}
\label{sec:deep_learning_model}

For the deep learning model, the model we chose was BanglaBERT (large), a BERT/ELECTRA model with 330M parameters pretrained on Bangla2B+ pretraining corpus. This model performed comparably to XLM-R (large) in NER (XLM-R (large) F1 score 78.39 vs. BanglaBERT (large) F1 score 79.20 on Semeval NER Bangla Dataset) while having fewer parameters (330M vs. 550M).

Furthermore, ELECTRA\cite{clark2020electra} is based using a generator-discriminator paradigm which makes it uniquely suitable for low compute environments. "Masked language modeling (MLM) methods, such as BERT, involve corrupting input by replacing some tokens with [MASK] and training a model to reconstruct the original tokens. These methods have been shown to produce good results when transferred to downstream NLP tasks, but they typically require large amounts of computational resources. An alternative approach, called replaced token detection, is proposed as a more sample-efficient method of pretraining. Instead of masking input, this method corrupts it by replacing some tokens with plausible alternatives sampled from a small generator network. A discriminative model is then trained to predict whether each token in the corrupted input was replaced by a generator sample or not. Experimentation demonstrates that this new pretraining task is more efficient than MLM because the task is defined over all input tokens rather than just the small subset that was masked out. This results in contextual representations that substantially outperform those learned by BERT, given the same model size, data, and computational resources. This advantage is particularly pronounced for small models. For example, the authors \cite{clark2020electra} showed that a model trained on one GPU for 4 days outperforms GPT, which was trained using 30x more compute, on the GLUE natural language understanding benchmark. The approach also works well at scale and performs comparably to RoBERTa and XLNet while using less than 1/4 of their compute and outperforms them when using the same amount of compute.

We started our preliminary tests on BanglaBERT (base) and BanglaBERT (large), tuning various hyperparameters. BanglaBERT (large) consistently outperformed its smaller counterpart, given enough training cycles to reach convergence. As such, BanglaBERT (large) was the final choice.

\section{Results}

\subsection{Feature Based Learning}

We iteratively added features that improved our score. The results for different combinations of the hand-crafted features can be found in Table 1.

\begin{table*}[t]
\centering
\begin{tabular}{|c|c|}
\hline
\textbf{Feature}                                                 & \textbf{F1 Score} \\ \hline
POS Tagger, Suffix                                               & 0.56              \\ \hline
POS Tagger, Suffix, k-Neighbor Words                             & 0.62              \\ \hline
POS Tagger, Suffix, k-Neighbor Words, Gazetteer Lists            & 0.689             \\ \hline
POS Tagger, Prefix, Suffix, k-Neighbor Words                     & 0.692             \\ \hline
\textbf{POS Tagger, Prefix, Suffix, k-Neighbor Words, k-means clustering} & \textbf{0.72}            \\ \hline
\end{tabular}
\caption{Iteratrive improvement of Feature Based Learning model}
\end{table*}

\subsection{Deep Learning Model}

\begin{table*}[t]
\centering
\begin{tabular}{|c|c|c|c|c|}
\hline
\textbf{Model} & \textbf{Batch Size} & \textbf{Max Seq Length} & \textbf{Epoch} & \textbf{F1 Score on dev.txt} \\ \hline
base & 16 & 128 & 3 & 0.73 \\ \hline
large & 16 & 128 & 3 & 0.77 \\ \hline
large & 32 & 64 & 3 & 0.76 \\ \hline
large & 16 & 128 & 6 & 0.78 \\ \hline
large & 32 & 64 & 6 & 0.79 \\ \hline
oversampled+large & 16 & 128 & 6 & 0.78 \\ \hline
SemEval2023data+large & 32 & 64 & 4 & 0.78 \\ \hline
SemEval2023data+weights+large & 32 & 64 & 4 & 0.74 \\ \hline
\textbf{SemEval2023data+large} & \textbf{32} & \textbf{64} & \textbf{6} & \textbf{0.79} \\ \hline
\end{tabular}
\caption{F1 Scores of different fine tuned Bangla BERT model}
\label{tab:F1_DL}
\end{table*}

\begin{table*}[t]
\centering
\begin{tabular}{|c|c|}
\hline
\textbf{Metric} & \textbf{Value}  \\ \hline
P-GRP & 0.8252427184466011 \\ \hline
R-GRP & 0.7203389830508469 \\ \hline
F1-GRP & 0.7692307692307188 \\ \hline
P-CORP & 0.7999999999999994 \\ \hline
R-CORP & 0.787401574803149 \\ \hline
F1-CORP & 0.793650793650743 \\ \hline
P-CW & 0.7563025210084028 \\ \hline
R-CW & 0.7499999999999993 \\ \hline
F1-CW & 0.7531380753137571 \\ \hline
P-PROD & 0.6783919597989946 \\ \hline
R-PROD & 0.7105263157894732 \\ \hline
F1-PROD & 0.6940874035989213 \\ \hline
P-LOC & 0.7909090909090902 \\ \hline
R-LOC & 0.8613861386138605 \\ \hline
F1-LOC & 0.8246445497629825 \\ \hline
P-PER & 0.9078947368421045 \\ \hline
R-PER & 0.9583333333333326 \\ \hline
F1-PER & 0.9324324324323818 \\ \hline
Precision & 0.7858910891089108 \\ \hline
Recall & 0.7937499999999998 \\ \hline
F1 & 0.7898009950248256 \\ \hline
\end{tabular}
\caption{Performance of Final Model}
\label{tab:F1_Final}
\end{table*}

Several strategies were explored to improve the robustness of our model. We first began by tuning maximum sequence lengths and batch sizes. The longest sentence in the training set was 34 words and allowing a maximum sequence length of 64 allows for dataset to fit into the model without truncation.
\section{Analysis}
\subsection{Feature Based Learning}
The problem with Feature Based Learning Models is that it is hard to scale and generalize for unseen datasets. Fetaures like Gazetteer Lists and Word Embedding based clusterings help generalize the model. We saw that feature based models struggles to identify CW tags. This is intuitive because these words depend on the context. And the handcrafted features are not able to identify that.  

\subsection{Deep Learning Model}
\subsubsection{Custom CrossEntropyLoss Weights}

Due the the data imbalance in the dataset, we tried a weighted CrossEntropyLoss function using the definition.
$$ w_i = 1 - \frac{n_i}{\sum_{j=1}^{N} n_j} $$
However, after training for 4 epochs, the F1 score was 0.74 which is lower than the unweighted model (0.78) while all other hyper-parameters remained the same. It it possible a different weight function could perform better, which is left as future work.

\subsubsection{Oversampling Certain Tags}
On un-augmented data, our selected model had the worse F1 scores on "CW" and "PROD" tags. We tried an oversampling policy of sampling sentences with "CW" and "PROD" two times. This resulted in no noticeable improvement, rather increased training time.

\subsubsection{Knowledge Base}
To overcome the problem mentioned for Creative Word (CW) tags, we wanted to leverage external knowledge. To do that we created a knowledge base from Wikipedia using Bangla Wikimedia dump. But the problem was that if we do knowledge retrieval and add context paragraphs with the sentence, the dataset becomes 6x large. And the max\_sequence\_length needs to be large too. Given our aws memory limit it is not possible to train the model with the aforementioned constraints. 

\section{Conclusion}
\label{sec:conclusion}

In this paper, we have presented two different approaches for Complex Named Entity Recognition (CNER) on BanglaCoNER dataset, a synthetic dataset that contains complex and compound entities in Bangla language. We have compared the performance of Conditional Random Fields (CRF) with various feature combinations and finetuning transformer based Deep Learning models such as BanglaBERT. Our results show that BanglaBERT (large) outperforms CRF by a significant margin, achieving an F1 Score of 0.79 on the validation set. This suggests that Deep Learning models can capture more linguistic patterns and nuances than traditional methods, especially for CNER task. Our study contributes to the advancement of NER research in Bangla language, which is underrepresented in natural language processing literature. We hope that our work will inspire more researchers to explore CNER task on other languages and datasets.

\bibliographystyle{IEEEbib}
\bibliography{refs}

\begin{thebibliography}{1}

\bibitem{ekbal2008named}
Asif Ekbal, Rejwanul Haque, and Sivaji Bandyopadhyay,
\newblock ``Named entity recognition in bengali: A conditional random field
  approach,''
\newblock in {\em Proceedings of the Third International Joint Conference on
  Natural Language Processing: Volume-II}, 2008.

\bibitem{mishra2016semi}
Shubhanshu Mishra and Jana Diesner,
\newblock ``Semi-supervised named entity recognition in noisy-text,''
\newblock in {\em Proceedings of the 2nd Workshop on Noisy User-generated Text
  (WNUT)}, 2016, pp. 203--212.

\bibitem{hasan-etal-2020-low}
Tahmid Hasan, Abhik Bhattacharjee, Kazi Samin, Masum Hasan, Madhusudan Basak,
  M.~Sohel Rahman, and Rifat Shahriyar,
\newblock ``Not low-resource anymore: Aligner ensembling, batch filtering, and
  new datasets for {B}engali-{E}nglish machine translation,''
\newblock in {\em Proceedings of the 2020 Conference on Empirical Methods in
  Natural Language Processing (EMNLP)}, Online, Nov. 2020, pp. 2612--2623,
  Association for Computational Linguistics.

\bibitem{clark2020electra}
Kevin Clark, Minh-Thang Luong, Quoc~V Le, and Christopher~D Manning,
\newblock ``Electra: Pre-training text encoders as discriminators rather than
  generators,''
\newblock {\em arXiv preprint arXiv:2003.10555}, 2020.

\end{thebibliography}

\end{document}